\documentclass{article}

\PassOptionsToPackage{numbers, compress}{natbib}

\usepackage[final]{neurips_2019}

\usepackage[utf8]{inputenc} 
\usepackage[T1]{fontenc}    
\usepackage{hyperref}       
\usepackage{url}            
\usepackage{booktabs}       
\usepackage{amsfonts}       
\usepackage{nicefrac}       
\usepackage{microtype}      

\usepackage{amsmath}
\DeclareMathOperator*{\argmax}{arg\,max}

\usepackage{amsmath, amssymb}
\usepackage{pgfplots}
\usepgfplotslibrary{fillbetween}
\usepackage{lipsum}
\usepackage{booktabs}
\usepackage{subcaption}
\usepackage{nameref}
\usepackage{algorithm}
\usepackage[noend]{algpseudocode}
\usepackage{wrapfig}
\usepackage{paralist}
\usepackage{titlesec}
\titlespacing\paragraph{0pt}{0pt plus 2pt minus 2pt}{2pt plus 2pt minus 2pt}

\newcommand{\toolName}{DRIFT}
\makeatletter
\def\blfootnote{\xdef\@thefnmark{}\@footnotetext}
\makeatother

\title{DRIFT: Deep Reinforcement Learning\\ for Functional Software Testing}

\author{%
  Luke Harries*, Rebekah Storan-Clarke*, Timothy Chapman, Swamy V. P. L. N. Nallamalli, \AND  Levent Ozgur, Shuktika Jain, Alex Leung, Steve Lim, Aaron Dietrich, \AND Jos\'e Miguel Hern\'andez-Lobato, Tom Ellis**, Cheng Zhang**, Kamil Ciosek** \\
}

\begin{document}

\maketitle

\blfootnote{$^*$ Equal contribution. Completed during the AI residency at Microsoft Research, Cambridge, UK. \\$^{**}$ Equal contribution from senior authors. All authors are affiliated with Microsoft.}

\vspace{-10pt}
\begin{abstract}
Efficient software testing is essential for productive software development and reliable user experiences. As human testing is inefficient and expensive, automated software testing is needed. In this work, we propose a Reinforcement Learning (RL) framework for functional software testing named \toolName{}. \toolName{} operates on the symbolic representation of the user interface. It uses Q-learning through Batch-RL and models the state-action value function with a Graph Neural Network. We apply \toolName{} to testing the Windows 10 operating system and show that \toolName{} can robustly trigger the desired software functionality in a fully automated manner. Our experiments test the ability to perform single and combined tasks across different applications, demonstrating that our framework can efficiently test software with a large range of testing objectives.
\end{abstract}
\section{Introduction}
Testing computer software is a crucial element of modern software engineering practice. The push towards continuous integration and continuous delivery (CI/CD) of software requires efficient testing to ensure the builds are stable \cite{stolberg_2009}. Otherwise, software delivery may be delayed or bugs may result in poor user experience.

While this is true for all software, tests are especially important for operating systems, where bugs could impair core system functions and introduce security vulnerabilities.

The most time-consuming part of software testing is testing through interactions with the Graphical User Interface (GUI), which was traditionally done manually. However, creating GUI tests manually is a time-consuming and expensive process. In particular, testing large numbers of interacting components takes many hours and small changes in the software can easily break many of these tests.  \cite{DBLP:journals/corr/abs-1812-11470}. As such, companies often outsource this testing to humans through quality assurance (QA) companies and/or beta users. 

Alternatively, large numbers of automated agents may be deployed to interact with the software. The current generation of these agents commonly act using a fixed random policy. This results in poorly-targeted tests, compared to those performed by a human user. To improve efficiency, heuristics have been added to random agents \cite{Machiry:2013:DIG:2491411.2491450}. However, a better automated solution for efficient software testing is clearly needed.

As system complexity increases, two features of software tests become critically important. First, since it is infeasible to manually specify a huge number of tests, they have to be fully automatic. Second, the testing framework needs to be \emph{sample-efficient}, meaning the ability to complete the tests with a reasonable amount of interactions. As discussed, existing approaches do not meet these requirements \cite{kochhar_thung_nagappan_zimmermann_lo_2015,mao_harman_jia_2016}.

In this work, we propose an efficient software testing framework addressing these requirements, by exploiting insights from Deep Reinforcement Learning \cite{SilverHuangEtAl16nature,mnih2015humanlevel,li2017deep}. Our goal is to train an RL agent which can perform efficient software testing with specified properties, such as testing specific functionalities with different coverage. We name our proposed framework DRIFT, which stands for Deep ReInforcement learning for Functional software-Testing.

\paragraph{Contributions} We design a novel Batch Reinforcement learning framework, DRIFT, for software testing. We use the tree-structured symbolic representation of the GUI as the state, modelling a generalizeable Q-function with Graph Neural Networks (GNN). We introduce a fully modular and automated setup to train agents to perform desired tasks. Additionally, the programmer can designate, in a language-agnostic way, which functionality should be tested. Afterwards, we evaluate DRIFT on the Windows 10 operating system, showing that trained agents can learn single tasks as well as multiple tasks with different coverage requirements. These agents outperform a random baseline by two orders of magnitude and can successfully generalize in settings where hash-based methods failed.

\section{Background}
To formally define our testing framework, we require several concepts that we will now introduce.

\paragraph{Markov Decision Process} We formalize the interaction between the agent and the environment as a family of Markov Decision Processes \cite{puterman2014markov} (MDP), indexed by the objective, $o$. An MDP is defined as a tuple $(S, A, T, R_{o}, \gamma, \bot_o)$, where $S$ is the set of states, $A$ is the set of actions, $T$ is the transition function, $R_o$ is the reward function corresponding to an objective $o$, $\gamma$ is the discount factor and $\bot_o$ is the set of absorbing states. The transition function $T(s_{t+1}, s_{t}, a_t) = \mathbb{P}(s'=s_{t+1} | s=s_{t},a=a_{t})$ models the transition to the next state given the current state and action. The reward function for a given objective $R_o(s_{t+1}, s_{t},a_{t}) = \mathbb{E}_{\tau}[r^o_{t} | s = s_{t},a=a_{t},s'=s_{t+1}]$ (we skip the superscript $o$ in the remainder of the paper where it is clear). Each step in the MDP can be described with a $\mathrm{transition} := (s_t, a_t, r_t, s_{t+1})$. An episode is a sequence of transitions until termination $\mathrm{episode} := [s_0, a_0, r_0, s_1, a_1, r_1 ...]$.

\paragraph{Reinforcement Learning}
In reinforcement learning, the agent is faced with a sequential decision-making problem. At each time step $t$,  the agent receives the state $s_t \in S$ from the environment. The agent has a policy $\pi$ which selects an action given the state $\pi(a|s) = P[A=a | S=s]$. The action is passed to the environment, and the state is updated using the transition function $T$. The environment then returns the new state $s_{t+1}$ and a scalar reward $r^o_{t+1} \in \mathbb{R}$ determined by the (objective-dependent) reward function. This interaction continues until a goal is achieved and thus the episode is completed. The task for the agent is to learn a policy $\pi$ which maximizes the total discounted reward $J^o_t = \mathbb{E}_{\tau}[\sum_{t} \gamma^{t} r^o_t]$ received from the environment where the discount $\gamma \in (0, 1]$ \cite{sutton_2018,abs-1801-10467}.

\paragraph{Trees and Graph Neural Networks}
A graph is a tuple $G = (V, E)$ containing a set of vertices $V$ and edges $E \subseteq \{(x, y) | (x, y) \in V^2 \land x \neq y\}$. A tree is an acyclic graph with a designated root node. Graph Neural Networks (GNNs) are differentiable parameterized functions $f: G \to \mathbb{R}^m$ where $v \in V, v \in \mathbb{R}^{n}$. Each layer in a GNN updates the representation of each node using its neighboring nodes \cite{scarselli2008graph,li2015gated,DBLP:journals/corr/abs-1812-08434}.

\paragraph{Off-Policy Reinforcement Learning and Batch Reinforcement Learning} 
Off-policy Reinforcement Learning is often deployed in settings where the policy collecting the experience is different from the policy being learned. This setting is known as off-policy reinforcement learning \cite{sutton1998introduction}. Batch Reinforcement Learning (Batch-RL) describes a subset of off-policy learning where a fixed set of transitions is used, with no further interaction with the environment \cite{lange_gabel_riedmiller_2012,DBLP:journals/corr/abs-1812-02900,DBLP:journals/corr/abs-1907-04543}

\section{Method}

We now describe the components of DRIFT. Abstractly, the framework works by exploiting the operating system API to interact with the software under test. It learns a policy which, given a symbolic representation of the interface, selects desired GUI interactions. 

In this section, we provide details of how this policy is obtained. We do this by first specifying software testing as a Markov Decision Process and then describing the training process that computes the policy for the MDP.

\begin{figure}
\begin{subfigure}[b]{.5\linewidth}
    \centering
    \begin{verbatim}
{
    "Identifier": "94d29a9543c9c...",
    "UIProperties": [
        {
            "AutomationID": "23423",
            "ClassName": "MainMenu"
            "ControlType": "Panel",
            "ProcessName": "StartMenu"
        }
    ],
    "Children": [...]
}
    \end{verbatim}
    \end{subfigure}
    \begin{subfigure}[b]{.5\linewidth}
    \centering
    \includegraphics[width=0.8 \linewidth]{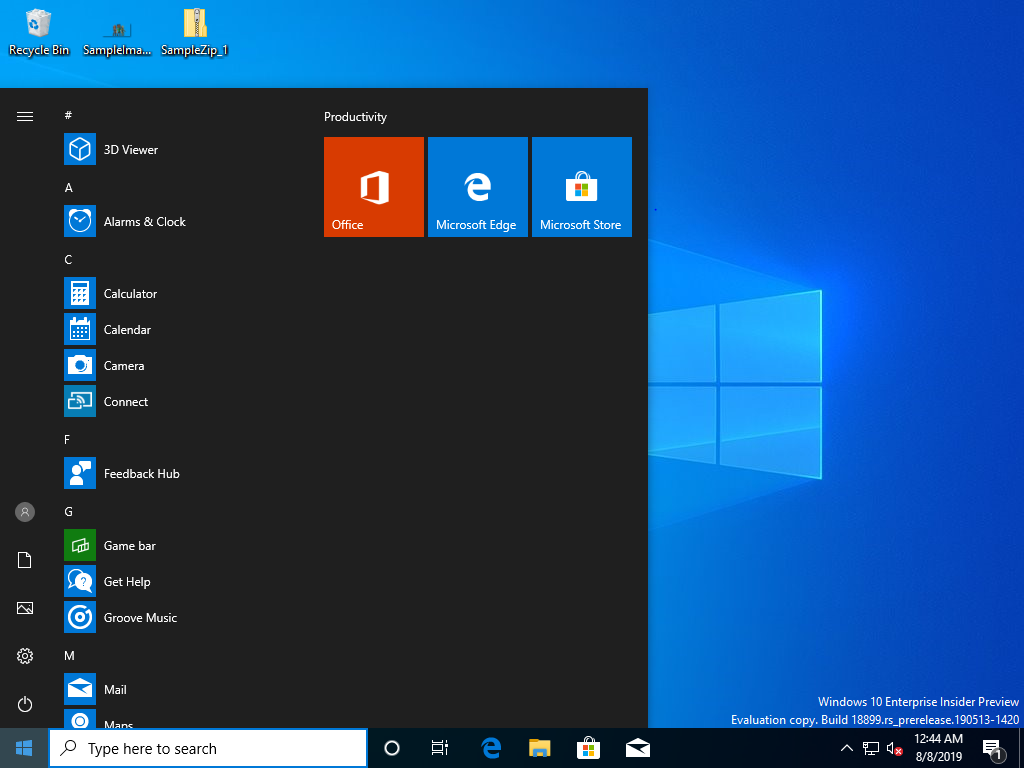}
    \end{subfigure}
    \caption{On left subfigure is the UITree for the StartMenu, shown by the large black rectangular box in the screenshot the right subfigure. On average, each state has 36 UI elements, although some  graphs have up to 800 UI elements.
    \vspace{-10pt}
    }
    \label{fig:startmenu}
\end{figure}

\subsection{The Testing MDP}

\paragraph{States} The state $s_t$ is a tree that corresponds to a hierarchy of GUI elements, known as a UITree. Each node in the UITree represents a UI element such as the "Start Menu" as shown in Figure \ref{fig:startmenu}. Each node has four properties: the \textit{AutomationID}, an optional string which is used for manual UI testing; the \textit{ClassName}, a string which determines the visual properties of the element; the \textit{ControlType}, an enum describing the type of element; and the \textit{ProcessName}, the name of the process which created the element. Additionally, the element may have children which correspond GUI elements that it contains. The root of the tree is the Desktop node.

The UITree is obtained using Microsoft UI Automation Tree Tool \cite{microsoft_2017} which is traditionally used for accessibility tools such as screen readers. Alternative tree representations of the interface may be used to test different systems, for example the Document Object Model (DOM) for testing websites.

An alternative to the UITree would be to use screenshots of the interface \cite{DBLP:journals/corr/abs-1901-02633}. However, using tree representations has several advantages over screenshots. Firstly, the features of each element are encoded directly into the node properties such as ClassName {Button, Label, ScrollBar, etc.} which eliminates the need to learn to visually recognize each element. Secondly, the UITrees are invariant to the location of the windows on the screen. Thirdly, the UITrees have a smaller memory footprint of a few KBs compared to several MBs for a screenshot. Fourthly, we have access to a very large number of historical trajectories where only the UITrees are stored. Combined, these features mean that a testing framework that makes use of UITrees is much more efficient and well worth the small amount of computation the API has to perform to obtain them.

\paragraph{Actions} The actions in the testing MDP reflect the possible interactions the user might have with the software such as clicking the mail icon to open to mail application. We extract the actions from the UITree. In particular, each possible action is a tuple consisting of a hash of the UIProperties of the node, known as the node identifier, and the action type, for example ("94d29a9543c9c...", "LeftClick"). As different windows may have a variable number of items the user can interact with, the number of actions that are available at each step is dynamic. The number of available actions varies from 2 to 842, with the mean being 248.

\paragraph{Reward Function}
 
For important systems, there often exists a list of key functions that should be verified as operational before a build is released. This process of verification is known as smoke testing \cite{memon_xie_2005}. In Windows 10, these key functions span many different apps. For example, adding a website to favorites in Edge or clicking "Add Bluetooth or other device" in System Settings.

We use several reward functions, one for each objective $o$ that our framework is trying to achieve. The reward function computes a scalar reward from a state-action pair. To know when a particular objective has indeed been achieved, we use an API function that triggers the generation of logs. The logs are converted into a scalar reward by counting the number of times the event related to a particular function has occurred. Our testing setup is fully modular, allowing the programmer to easily mark, in a language-agnostic way, which functionality should be tested. The agent adapts both to the objective and to any changes in the software that might have changed the best path towards the existing objectives.

\subsection{Training}

\paragraph{The Simulator}

We developed a training environment by interfacing with the Windows operating system. We followed OpenAI's gym specification \cite{1606.01540}. However, since GUI events are processed in real-time, the time required to complete each step (such as opening Outlook) can be up to several seconds. This is very different from step lengths of several milliseconds, typical of most typical RL environments. This, combined with the often complex sequence of actions needed to perform a particular task, impairs the ability to directly learn from the simulator in a reasonable length of time. Therefore, we trained our agent using a large cache of historical data. The simulator was then only used to evaluate the agents. The simulator was reset when the desired task was achieved.

\paragraph{Historical Data}
To overcome the limitations of the relatively slow simulator, we learn solely from historical data. In our case, the historical data are the episodes collected by random agents during previous testing runs. We query the historical data for the particular episodes where a particular objective is achieved, helping to alleviate the problem of sparse rewards. Our testing framework is generic with respect to the source of historical data. For alternative systems under test, interactions recorded from real users could be used as well.

\paragraph{Q-learning}

For training, we use a variant of Q-learning known as DQN \cite{sutton1998introduction,mnih2015humanlevel}. As Q-learning is an off-policy algorithm, the training data does not have to be collected from the policy being trained. In Q-learning, we learn the Q-function which allows us to estimate the expected cumulative reward of a state-action pair under the optimal policy $\pi^\star$.
By default, the policy $\pi$ of the agent is \textit{greedy} with respect to the current estimate of the Q-function, selecting the action with the highest expected reward $\pi (s) = \argmax_{a \in A} Q_\pi(s, a). $
Q-learning iteratively performs the update $Q'_\pi(s, a) \gets r + \max_{a \in A}Q(s', a)$ for tuples $(s,a,r,s')$.

Due to the large possible number of combinations of state-action pairs, and to achieve the ability to generalize across state-action pairs, we approximate the function using a neural network and optimize the network parameters $\mathbf{\theta}$ by using stochastic gradient descent to minimize the loss function $L$. The stochastic gradient descent update is shown below where $\beta$ represents the learning rate. In practice, the learning rate is dynamically set using the Adam optimiser \cite{Adam} with an initial value of $1 \times 10^{-2}$.

The combination of approximation (our use of neural networks), bootstrapping, and off-policy learning results in instability during training known as the deadly triad \cite{sutton1998introduction}. We use an experience replay buffer and dual Q networks to aid stability \cite{mnih2015humanlevel}. The loss function $L$ is based on \cite{mnih2015humanlevel}, $D$ represents the dataset of transitions $(s, a, r, s')$ contained in the experience replay, $\mathbf{\theta}$ represents the parameters of the policy network, and $\mathbf{\theta}^{-}$ represents the values of the target network
\begin{equation}
L(\mathbf{\theta}_{i}) = \mathbb{E}_{(s, a, r, s') \sim D}[(r + \gamma \max_{a' \in A} Q(s', a'; \mathbf{\theta}_i^{-}) - Q_\pi(s, a; \mathbf{\theta}_i))^2]
\end{equation}
An additional parameter $\eta \in \mathbb{R}^{+}$ determines the frequency at which the target network parameters are updated using the policy network parameters. $\eta$ and $\gamma$ are chosen using a hyperparameter sweep.
Algorithm \ref{algo:batch-rl} gives an overview of our training procedure.

\paragraph{Modeling the state using Graph Neural Networks}

The state is initially returned from the environment as a UITree. We first convert the nodes of the UITree to vector form and then apply the graph neural network.

For a graph with $n \in \mathbb{N}^+$ nodes, we set the representation of each node as the concatenation of the one hot encoding of the node's \textit{UIProperties}. Specifically, we use a variant of one hot encoding where the rarely seen/unseen properties are grouped into an "Other" value. This ensures that the network can handle unseen values at evaluation time. We represent each node embedding $\mathbf{v} \in {\{0,1\}}^z$. We define the vectorized action $\mathbf{a} \in (\mathbf{a}_e, \mathbf{a}_i)$ as the concatenation of the one hot encoding of the action type $\mathbf{a}_e \{\mathrm{LeftClick}, \mathrm{RightClick}, ...\}$ and $\mathbf{a}_i$ the one hot encoding of the node index $[0..n]$. 

We use a graph neural network (GNN) to approximate the action-value function where $s = (V_s, E_s)$ and $Q(s, a) = \textrm{GNN}(V_s, E_s) \cdot \mathbf{a}.$
The GNN is applied to the state and outputs a matrix containing a vector representation for each node. The expected cumulative reward for performing each of the corresponding actions, followed by the agent acting greedily, is calculated by performing a dot product with the vectorized action.
Here, we want a network architecture which takes into account the hierarchy within the tree and is able to differentiate between nodes with identical properties but different locations within the tree. We chose the GNN architecture as it is specially designed to operate on graph structures \cite{DBLP:journals/corr/abs-1812-08434}. 

\begin{wrapfigure}{l}{0.5\textwidth}
\begin{minipage}{0.5\textwidth}
\begin{algorithm}[H]
\caption{\toolName{} Batch-RL}\label{euclid}
\begin{algorithmic}
\State \textbf{Input:} The desired objective \textit{o}, the application processes relevant to the reward \textit{process}, the historical data \textit{D}
\State \textbf{Output:} Trained GNN \textit{net} \\
\State $\textit{transitions} \gets \textit{list}()$\\

 \State \# Extract relevant transitions
\For{\textit{episode} in \textit{D}}
    \If {\textit{episode\_meets\_objective(episode, o)}}
        \State $\textit{cropped\_episode} \gets \textit{crop(episode, o)}$
        \State \textit{transitions += cropped\_episode}
    \EndIf
\EndFor \\

\State \# Vectorize the states and actions
\For{\textit{t} in \textit{transitions}} 
    \For{\textit{s in ['state', 'next\_state']}}
    \State $\textit{t[s]} \gets \textit{get\_process(t[s], process)}$
    \State $\textit{t[s]} \gets \textit{vectorise(t[s])}$
    \EndFor
    \State $\textit{t['r']} \gets$ \textit{get\_reward(o, t)}
    \State $\textit{t['a']} \gets \textit{vectorise\_action(t['a'])}$
\EndFor \\

\State $\textit{net} \gets \textit{GNN()}$ 
\State $\textit{target\_net} \gets \textit{net.copy()}$
\State $\textit{update\_frequency} \gets 100$
\State $\textit{steps} \gets 0$
\State \# Train the GNN
\For{\textit{batch} in \textit{DataLoader(transitions)}}
    \State $\textit{net} \gets \textit{train(batch, net, target\_net)}$
    \If{$\textit{update\_frequency mod steps} = 0$}
        \State $\textit{target\_net} \gets \textit{net.copy()}$
    \EndIf
\EndFor
\end{algorithmic}
\label{algo:batch-rl}
\end{algorithm}
\end{minipage}
\end{wrapfigure}

At each time step or each layer, the GNN updates the representation of each node based on the current node representation and its neighboring nodes. We chose GNN architecture by using a toy supervised learning task (classifying whether the node ClassName is Button). We selected the Graph Attention Network (GAT) \cite{velickovic2018graph} as it achieved the best performance. GAT is a convolution style architecture which uses self-attention to determine the relative weighting of the neighboring nodes. Each layer of the GAT updates each node representation
\begin{equation}
    \mathbf{v'}_{i} = \alpha_{i,i} \mathbf{v}_{i} +  \sum_{j \in \mathcal{N}(i)}{\alpha_{i,j} \mathbf{v}_{j}}.
\end{equation}
where $\alpha_{i,j}$ represents the self-attention with $i$ being the index of the query node and $j$ being the index of the key node.

We implemented the GAT architecture using PyTorch Geometric \cite{FeyLenssen2019}. Our final architecture consisted of two GAT layers separated by a Rectified Linear Unit (ReLU) \cite{relu}. The first layer has 1080 input channels (the length of the vectorized node representation) and 80 outputs channels with 8 self-attention heads. The second layer outputs 1 channel (the number of action types) and 1 self-attention head. Only 1 action type was used as all the key functions could be performed with only the \textit{LeftClick}.

\begin{figure*}[t]
    \centering
    \includegraphics[width=0.3\linewidth]{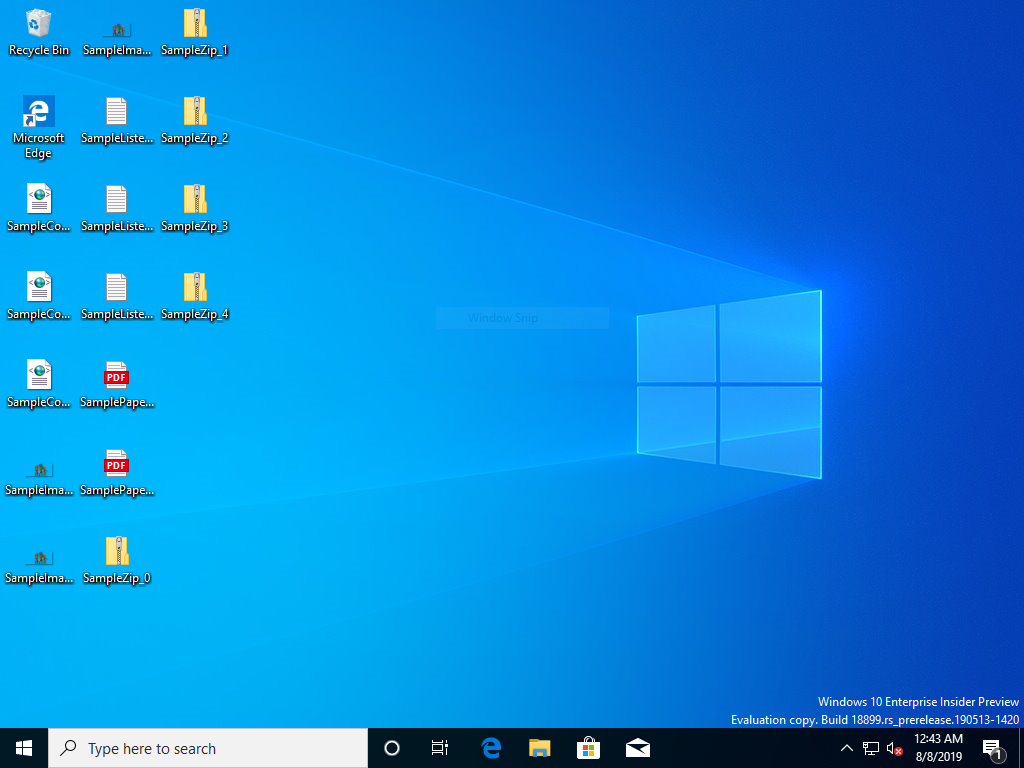} 
    \hspace{15pt}
    \includegraphics[width=0.3\linewidth]{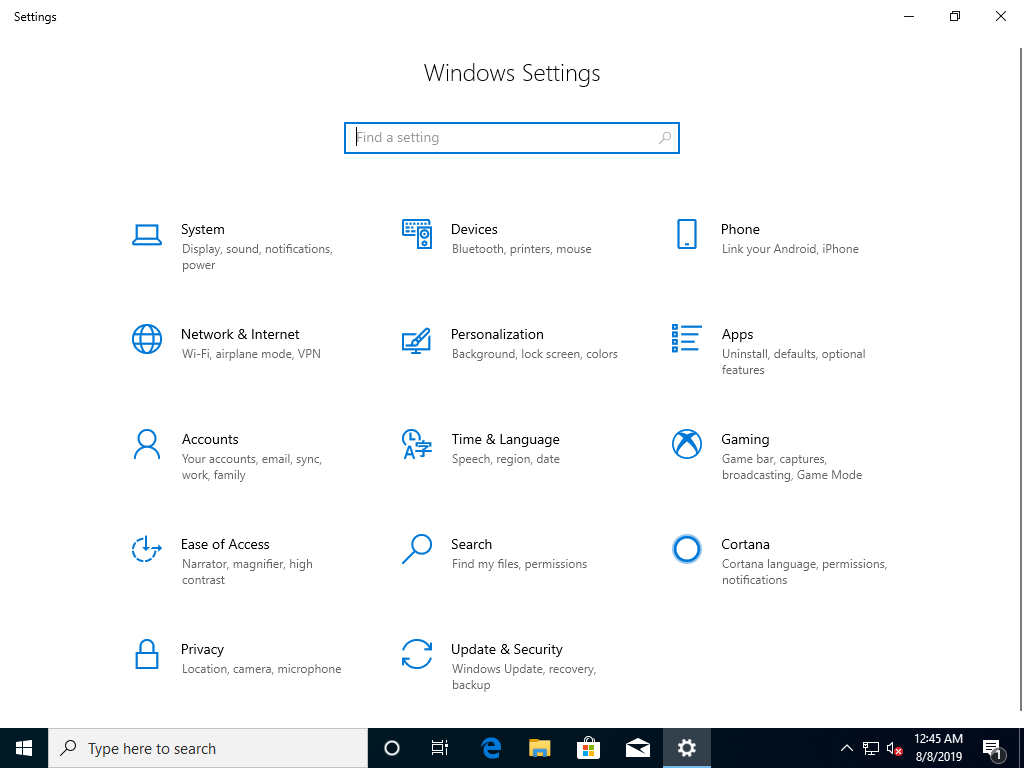} 
    \hspace{15pt}
    \includegraphics[width=0.3\linewidth]{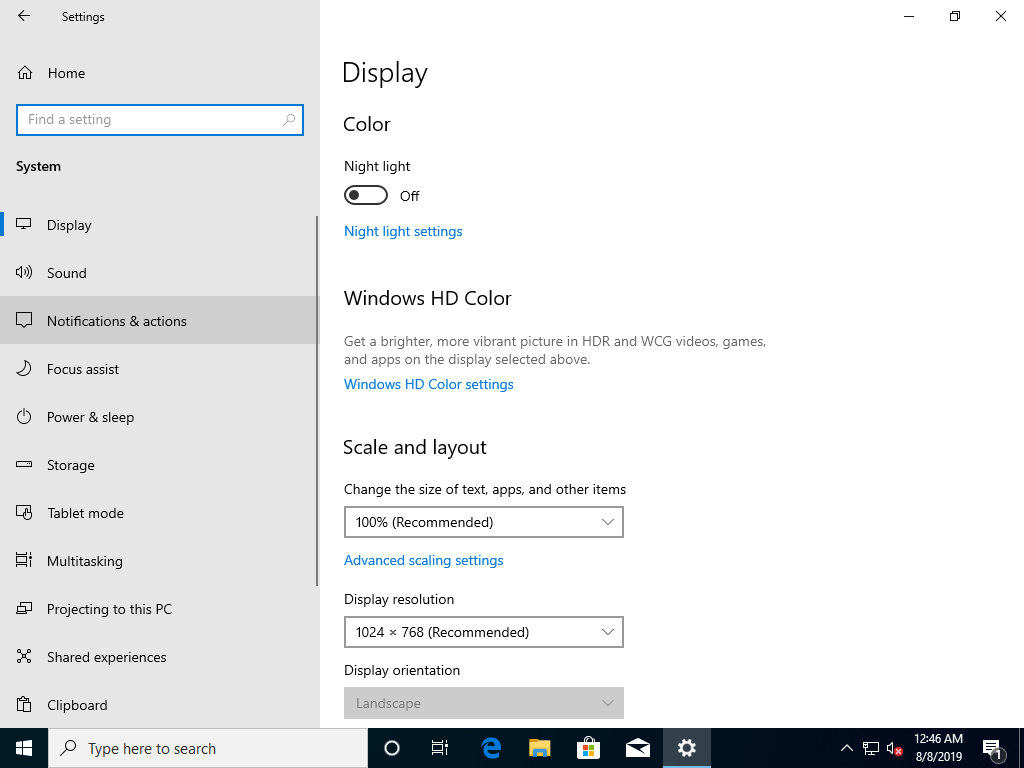}
    \caption{States seen whilst navigating to the notifications panel. At the start of each episode, the System Settings process is started by the simulator..
    To achieve the reward the agent must select the "System" element followed by the "Notifications Panel" element.
    \vspace{-10pt}
    }
    \label{fig:notificationspanel}
\end{figure*}

\paragraph{Efficiency vs Coverage}

We introduce two variants of \toolName{}: \toolName{}-Greedy and \toolName{}-Sampler. \toolName{}-Sampler focuses on performing one given objective most efficiently. It uses a greedy policy, which selects the action with the highest expected reward $\pi(s, a) = \argmax_{a \in A}Q(s,a)$. 
However, when testing software, we may also want to maximize state coverage in addition to achieving the desired objective. To do this, we use \toolName-Sampler, which has a stochastic policy $\pi(s, a) = a \sim \mathrm{Categorical}(\mathrm{Softmax}([Q(s, a)/m | a \in A]))$ where the temperature parameter $m \in (0,\infty]$. Smaller temperature values bias the sampled actions towards the action with the largest predicted reward while higher temperature values bias the sampled actions toward a uniform distribution across actions.

\paragraph{Multiple Tasks}

To support multiple tasks, \toolName{} is trained train on episodes where the tasks are   performed individually. Since the rewards are normalized across tasks, this gives a combined policy that targets all tasks. During training, we monitor performance on each task. During testing, we deploy the multi-objective agent using \toolName-Sampler so that the agent explores more efficiently and is more robust to imperfections in the learned policy.

\section{Experiments}

\begin{wrapfigure}{r}{0.5\textwidth}
\centering
    \scalebox{0.6}{
    \begin{tikzpicture}
        \begin{axis}[
            xlabel={Training steps}, 
            ylabel={Total reward over $1000$ evaluation steps},
            legend pos=south east
        ]
        \addplot [blue] table [x=x, y=y, col sep=comma] {single_task.csv};
        \addlegendentry{\toolName-Greedy}
        
        \addplot [red, dashed] table [x=a, y=b, col sep=comma] {single_task_random.csv};
        \addlegendentry{Random Agent}

        \addplot [name path=upper,draw=none] table[x=x,y expr=\thisrow{y}+\thisrow{err}, col sep=comma] {single_task.csv};
        \addplot [name path=lower,draw=none] table[x=x,y expr=\thisrow{y}-\thisrow{err}, col sep=comma] {single_task.csv};
        \addplot [fill=blue!10] fill between[of=upper and lower];

        \end{axis}
    \end{tikzpicture}
    }
    \caption{The cumulative reward achieved by an agent on a single task for a given amount of training. The task is to navigate to the notifications page within the settings app for which it is given a reward of 1.
    The shaded area represents the standard deviation of the \toolName{}-Greedy agent over 5 folds and 4 seeds.}
    \label{fig:single_task}
\end{wrapfigure}
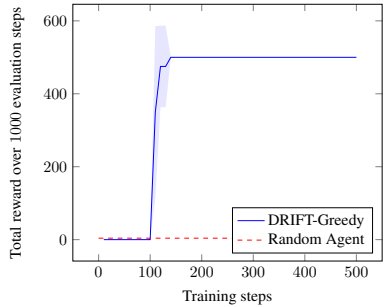

\begin{figure*}[t]
    \centering
    \begin{tabular}{c|c|c|c}
    \toprule
     Task & DRIFT-Greedy & Q-hash Agent & Random Agent \\
      & (Train, Eval) & (Train, Eval) & (Evaluation) \\
     \midrule
     Navigate to notifications panel (Settings) & $2 \pm 0$, \textbf{$\boldsymbol{2} \pm 0$} & $2 \pm 0$, Fail & $443 \pm 1235$\\
     Add Bluetooth or other device (Settings) & $2 \pm 0$, \textbf{$\boldsymbol{2} \pm 0$} & $2 \pm 0$,  Fail & $852 \pm 1681$\\
     Add a page to favorites (Edge) & $2 \pm 0$, \textbf{$\boldsymbol{2} \pm 0$} & $2 \pm 0$, Fail & $444 \pm 1372$ \\
    \bottomrule 
    \end{tabular}
    \caption{The number of steps required for each agent to complete a specific task on a simulator. For \toolName{}-Greedy and Q-hash, the evaluation was performed using 1000 evaluation steps by the agent in the simulator. The result shown is the mean and standard deviation of 5-fold cross-validation with 4 random seeds. The number of steps for the Random Agent is calculated using the mean historical data and includes the standard deviation. \toolName{}-Greedy was able to successfully learn to achieve all the tasks following the optimal route.
    Although the random agent was able to achieve all the tasks, it was very \textit{sample-inefficient}, requiring a large number of steps. The Q-hash based agent failed to achieve any of the tasks on the unseen evaluation environments as it was unable to generalize.
    }
    \label{fig:single_rewards}
\end{figure*}

We use \toolName{} to perform a variety of different software tests within Windows 10. In particular, we verify key functionality of the System Settings App and the Edge browser. \toolName{}-Greedy finds the most efficient path to the objective, outperforming the random agent baseline by two orders of magnitude, and consistently achieving the reward, unlike the Q-hash baseline which failed to generalize. We show that the temperature parameter of \toolName{}-Sampler can be varied to effectively trade-off between efficiency at completing the task and coverage of the possible states. Additionally, we demonstrate that \toolName{}-Sampler can learn to accomplish multiple testing objectives.

\paragraph{Experimental setup} The agents were trained on historical data as described in the Methods section. For each reward, 20 episodes of training data were used, each with an average of 44 transitions. We used a batch size of 128 transitions, where each batch is known as a step. We used random search to determine which hyperparameters were able to most efficiently learn to navigate to the notifications page. The best frequency of the target network updates was found to be 100 and the optimal discount factor $ \gamma $ was set to 0.1. All experiments are repeated using 5-fold cross-validation, holding out a subset of the historical data used for training, with 4 different seeds. We implemented the algorithm using PyTorch and PyTorch Geometric \cite{paszke2017automatic,FeyLenssen2019}

As the naive agent required a huge number of samples to complete the tests it was very expensive to evaluate, taking on average 443 steps to 852 steps with a very large standard deviation to achieve one desired goal state. Rather than let it timeout, we performed the evaluation off-policy. 

We use an additional baseline, which we call Q-hash, where the representation of a state-action pair was obtained by hashing. Q-hash learned the correct path on the training data, however, it failed on both held-out data and the evaluation simulator. This was caused by the fact that slight variations in the state greatly change the hash representation, which meant the algorithm was unable to generalize. Although method was inspired by \cite{vos_2015}, we do not use the same feature engineering, or a process of selecting stable elements, which is the reason our Q-hash agent was unable to generalize.

\subsection{Solving a specific task}

We evaluated the ability of \toolName{}-Greedy to perform tasks by measuring the number of times a particular task is achieved in 1000 steps. For example, the "Navigate to notifications panel" task is shown in Figure \ref{fig:notificationspanel}. Figure \ref{fig:single_task} demonstrates that all runs of DRIFT, regardless of which seed or folds of data were used for training, were able to learn the most efficient route to the notifications panel (2 steps). In contrast, by analyzing the historical data we can see that the random agent took on average 443 steps with a very large standard deviation. The Q-hash approach failed to perform the task due to an inability to generalize to slight differences in the state of the simulator compared with the training data. Results for a larger number of different tasks are shown in Figure \ref{fig:single_rewards}.

\subsection{Trading off efficiency and state coverage}

\begin{figure*}[t]
    \centering
    \scalebox{0.6}{
    \begin{tikzpicture}
        \begin{axis}[
            xlabel={Temperature},
            xmode=log,
            ylabel={Total reward over $1000$ evaluation steps},
            legend pos=north east
        ]
        \addplot [blue] table [x=x, y=y, col sep=comma] {temperature.csv};
        \addlegendentry{\toolName-Sampler}
        
        \addplot [red, dashed] table [x=a, y=b, col sep=comma]{temperaturerandom.csv};
        \addlegendentry{Random Agent}

        \addplot [mark=none, name path=upper,draw=none] table[x=x,y expr=\thisrow{y}+\thisrow{err}, col sep=comma] {temperature.csv};
        \addplot [name path=lower,draw=none] table[x=x,y expr=\thisrow{y}-\thisrow{err}, col sep=comma] {temperature.csv};
        \addplot [fill=blue!10] fill between[of=upper and lower];
        
        \addplot [mark=none, name path=upper,draw=none] table[x=a,y expr=\thisrow{b}+\thisrow{std}, col sep=comma] {temperaturerandom.csv};
        \addplot [name path=lower,draw=none] table[x=a,y expr=\thisrow{b}-\thisrow{std}, col sep=comma] {temperaturerandom.csv};
        \addplot [fill=red!10] fill between[of=upper and lower];

        \end{axis}
    \end{tikzpicture}
    }
    \hspace{10pt}
    \scalebox{0.6}{
    \begin{tikzpicture}
        \begin{axis}[
            xlabel={Temperature},
            xmode=log,
            ylabel={Number of unique states seen},
            legend pos=north west
        ]
        \addplot [blue] table [mark=none, x=x, y=y, col sep=comma] {pages_seen.csv};
        \addlegendentry{\toolName-Sampler}
        
        \addplot [red, dashed] table [mark=none, x=x, y=y, col sep=comma] {randompagesseen.csv};
        \addlegendentry{Random Agent}

        \addplot [mark=none, name path=upper,draw=none] table[x=x,y expr=\thisrow{y}+\thisrow{std}, col sep=comma] {pages_seen.csv};
        \addplot [name path=lower,draw=none] table[x=x,y expr=\thisrow{y}-\thisrow{std}, col sep=comma] {pages_seen.csv};
        \addplot [fill=blue!10] fill between[of=upper and lower];
        
        \addplot [mark=none, name path=upper,draw=none] table[x=x,y expr=\thisrow{y}+\thisrow{std}, col sep=comma] {randompagesseen.csv};
        \addplot [name path=lower,draw=none] table[x=x,y expr=\thisrow{y}-\thisrow{std}, col sep=comma] {randompagesseen.csv};
        \addplot [fill=red,opacity=0.2] fill between[of=upper and lower];

        \end{axis}
    \end{tikzpicture}
    }
    \caption{Number of rewards achieved and unique pages visited given different temperature. The task is to navigate to the notifications panel within the settings app.
    As the temperate increases, the distribution from which the actions are sampled becomes more uniform and so more states are visited. The shading represents the standard deviation across 5 folds and 4 seeds.
    }\label{fig:temperaturevaries}

\end{figure*}
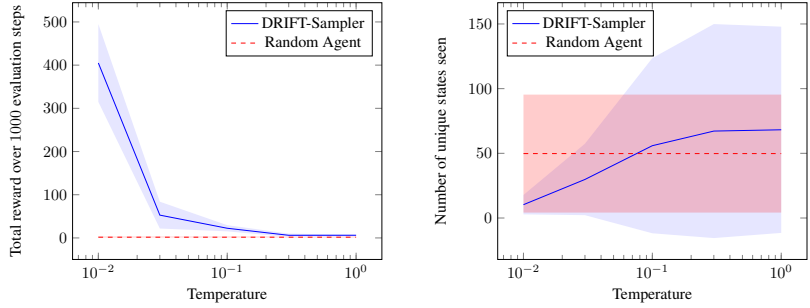
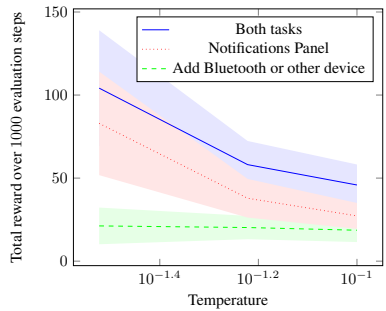
\begin{wrapfigure}{l}{0.5\textwidth}
\centering
    \scalebox{0.6}{
    \begin{tikzpicture}
        \begin{axis}[
            xlabel={Temperature},
            xmode=log,
            ylabel={Total reward over $1000$ evaluation steps},
            legend pos=north east
        ]
        
        \addplot [blue] table [x=temperature, y=TotalMean, col sep=comma] {multiplerewards.csv};
        \addlegendentry{Both tasks}
        
        \addplot [red, dotted] table [x=temperature, y=NotificationMean, col sep=comma] {multiplerewards.csv};
        \addlegendentry{Notifications Panel}
        
        \addplot [green, dashed] table [x=temperature, y=DeviceMean, col sep=comma] {multiplerewards.csv};
        \addlegendentry{Add Bluetooth or other device}

        \addplot [mark=none, name path=upper,draw=none] table[x=temperature,y expr=\thisrow{TotalMean}+\thisrow{TotalStd}, col sep=comma] {multiplerewards.csv};
        \addplot [name path=lower,draw=none] table[x=temperature,y expr=\thisrow{TotalMean}-\thisrow{TotalStd}, col sep=comma] {multiplerewards.csv};
        \addplot [fill=blue!10] fill between[of=upper and lower];

        \addplot [mark=none, name path=upper,draw=none] table[x=temperature,y expr=\thisrow{DeviceMean}+\thisrow{DeviceStd}, col sep=comma] {multiplerewards.csv};
        \addplot [name path=lower,draw=none] table[x=temperature,y expr=\thisrow{DeviceMean}-\thisrow{DeviceStd}, col sep=comma] {multiplerewards.csv};
        \addplot [fill=green!10] fill between[of=upper and lower];

        \addplot [mark=none, name path=upper,draw=none] table[x=temperature,y expr=\thisrow{NotificationMean}+\thisrow{NotificationStd}, col sep=comma] {multiplerewards.csv};
        \addplot [name path=lower,draw=none] table[x=temperature,y expr=\thisrow{NotificationMean}-\thisrow{NotificationStd}, col sep=comma] {multiplerewards.csv};
        \addplot [fill=red!10] fill between[of=upper and lower];
        
        \end{axis}
    \end{tikzpicture}
    }
    \caption{The cumulative reward achieved by \toolName{}-Sampler on multiple tasks given different temperature. The agent is given a reward of 1 for each of the tasks: firstly, navigating to the notifications panel;
     secondly, navigating to and clicking the button "Add Bluetooth or other device". 
    \vspace{-20pt}
    }\label{fig:multiplerewards}
\end{wrapfigure}
 An additional consideration when performing automated software testing is covering a large number of possible states. We evaluate \toolName-Sampler's ability to trade-off coverage and ability to achieve the desired reward by investigating the effects of the temperature parameter. The left panel of Figure \ref{fig:temperaturevaries} shows that as we increase the temperature the number of steps required to achieve the reward decreases. The right panel shows that as the temperature increases the number of unique states visited increases as well. Together, these panels show that the temperature can be used to effectively trade-off between task efficiency and state coverage given a trained value function.

\subsection{Multiple Testing Objectives}

We evaluated our testing framework where there were multiple functions to test, as is often the case in practice. To do this, we trained a single agent on a mixture of episodes containing a mixture of several reward signals.  

We trained the agent to perform two tasks within the Settings application. The first task was to navigate to the notifications panel, as shown in Figure \ref{fig:notificationspanel}, requiring two steps. The second task was to navigate to the "Devices" page and then click "Add Bluetooth or other device", similarly requiring two steps. A reward of 1 was given for completing either task.

As our agent has no memory, learning to achieve the tasks sequentially was not possible. Instead, we used the \toolName{}-Sampler so that at each step the agent would sample from the likely actions and, hopefully, on different runs, randomly choose between the different objectives. As shown in Figure \ref{fig:multiplerewards} the agent was able to solve both tasks at least some of the time over 1000 evaluation steps. However, the "Add Bluetooth or other device" task was completed less often. On average, it was accomplished 21 times using a temperature of 0.03 and 18 times using a temperature of 0.1. This was due to the fact that some of the training episodes for the Bluetooth task also contained successful solutions to the Notifications panel task, while the opposite was not the case, biasing the agent to one of the tasks.

\section{Related Work}

Our framework has the same underlying objective as other tools for automated software testing --- using the GUI to find the kind of bugs that are easily reachable by the end-user \cite{android_developers,vos_2015,mao_harman_jia_2016}. 
However, many of these tools use a policy without a feedback loop, simply generating a sequence of inputs from a fixed distribution. For example, Monkey uses a random policy to test android apps with the action space being a combination of UI interactions and system events such as turning airplane mode off and on \cite{android_developers}. GUITest is a Java application which uses a fixed random policy to test MacOSX apps \cite{bauersfeld_vos_2012}, similarly using the accessibility interface of the operating system. Such fixed policies are often combined with heuristics. For example, the DynoDroid policy is biased towards least recently used actions \cite{Machiry:2013:DIG:2491411.2491450}. The common limitation of these tools is that they require large amounts of time to complete the tests.

Several other tools have been produced to deal with this limitation. Sapienz uses genetic algorithms to optimize the sequences generated by the random policy \cite{mao_harman_jia_2016}. Sapienz learns to perform a sequence of events and motifs (hand-crafted sequence of events) depending on the current screenshot. Humanoid \cite{DBLP:journals/corr/abs-1901-02633} learns to generate human-like actions by using a convolutional neural network to learn a mapping from the screenshot to the actions selected by end-users.
It has been theorized that the improved coverage compared to other tools was the result of learning to prioritize more critical UI elements.

The system \cite{Bauersfeld2012ARL} proposed a Q-learning based approach to finding bugs within MacOSX applications. They suggested using a Q-table, where a Q-value is learned for each state-action pair. To do this, they propose hand-engineering the representation of the state and actions which can then be used to lookup the Q-table for the corresponding Q-value. This approach was developed by \cite{vos_2015} resulting in the "Testar" tool, where the state and action representations are created by their respective hashes on hand-selected, application-specific stable elements. This paper inspired our Q-hash baseline. However, we found that our model, which lacked feature-engineering of stable elements, was unable to generalize to any small variations in the state and actions, suggesting that this method is not robust.

\section{Conclusion}
We propose \toolName{}, a novel efficient software testing framework. We first formalize the software testing task as an MDP and solve it using deep RL. Our agent operates on a symbolic representation of the GUI and uses a graph neural network to model the state-action value function. To amortize the cost of data collection, it is trained from existing data using the batch-RL paradigm. We demonstrate our framework on several testing tasks on the Windows 10 platform. Our agents outperform the baseline, a fuzzing tool with a policy independent of the state, by two orders of magnitude. Moreover, we compare to a simpler Q-learning approach, which represents the state-action tuples using a hash function, that failed to generalize. We introduce a sampling agent and demonstrate that we can trade-off efficiency and coverage in addition to learning to perform multiple tasks.
In future work, we hope to improve exploration by incorporating agent memory, which would allow more complex tests, where the current action depends on the whole history of interactions between agent and environment rather than just the current state.
\clearpage

\bibliographystyle{unsrt}
\bibliography{references}

\end{document}